\documentclass[runningheads]{llncs}

\usepackage{eccv}

\usepackage{eccvabbrv}

\usepackage{graphicx}

\usepackage{booktabs}
\usepackage{tabularx}
\usepackage{makecell}   %
\usepackage{comment}
\usepackage{pifont}

\usepackage{siunitx}
\newcommand{\cmark}{\ding{51}} %
\newcommand{\xmark}{\ding{55}} %

\newcolumntype{L}[1]{>{\raggedright\arraybackslash}p{#1}}
\newcolumntype{C}[1]{>{\centering\arraybackslash}p{#1}}
\newcommand{\hdr}[1]{\shortstack{#1}}

\newcolumntype{Y}{>{\raggedright\arraybackslash}X}

\usepackage{hyperref}

\begin{document}

\title{GOLD-BEV: GrOund and aeriaL Data for Dense Semantic BEV Mapping of Dynamic Scenes} 
\titlerunning{GOLD-BEV}

\author{Joshua Niemeijer\inst{1} \and
Alaa Eddine Ben Zekri\inst{1} \and
Reza Bahmanyar\inst{1} \and 
Philipp M. Schmälzle\inst{1} \and 
Houda Chaabouni-Chouayakh\inst{2} \and 
Franz Kurz\inst{1}
}

\authorrunning{J. Niemeijer, A. E. Ben Zekri, R. Bahmanyar, P. M. Schmälzle et al.}

\institute{
German Aerospace Center (DLR)\\
\email{\{joshua.niemeijer, alaa.benzekri, reza.bahmanyar, philipp.schmaelzle, franz.kurz\}@dlr.de}
\and
Digital Research Center of Sfax \\
\email{houda.chaabouni@crns.rnrt.tn}
}

\maketitle

\begin{abstract}
Understanding road scenes in a geometrically consistent, scene-centric representation is crucial for planning and mapping. We present \textbf{GOLD-BEV}, a framework that learns dense bird’s-eye-view (BEV) semantic environment maps---including dynamic agents---from ego-centric sensors, using \textbf{time-synchronized aerial imagery} as supervision \emph{only during training}. BEV-aligned aerial crops provide an intuitive target space, enabling dense semantic annotation with minimal manual effort and avoiding the ambiguity of ego-only BEV labeling. Crucially, strict aerial--ground synchronization allows overhead observations to supervise moving traffic participants and mitigates the temporal inconsistencies inherent to non-synchronized overhead sources. To obtain scalable dense targets, we generate BEV pseudo-labels using domain-adapted aerial teachers, and jointly train BEV segmentation with optional pseudo-aerial BEV reconstruction for interpretability. Finally, we extend beyond aerial coverage by learning to synthesize pseudo-aerial BEV images from ego sensors, which support lightweight human annotation and uncertainty-aware pseudo-labeling on unlabeled drives. 
\end{abstract}

\begin{figure*}[t]
  \centering
  \includegraphics[width=0.6
  \textwidth]{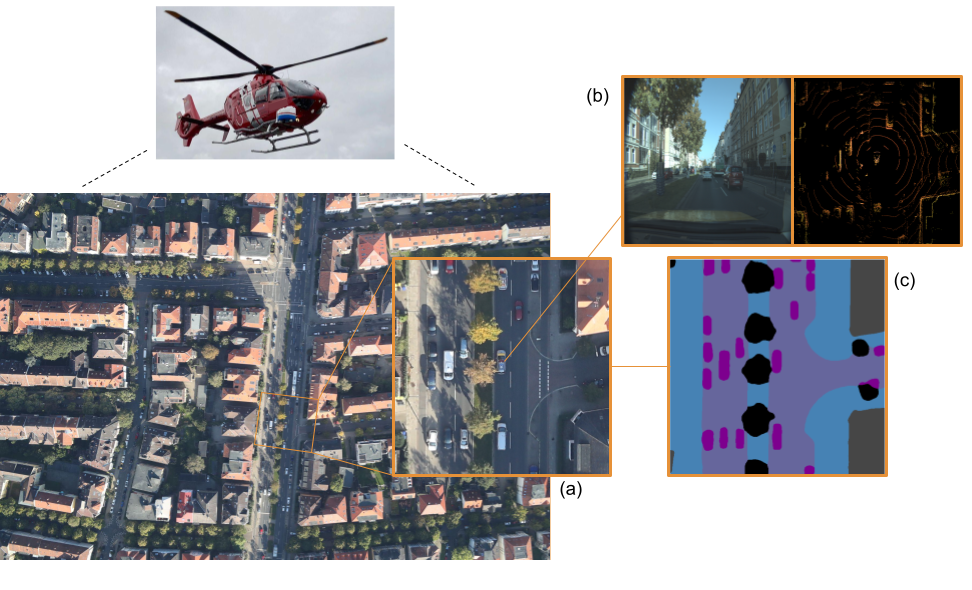} %
  \caption{Cross-view aerial supervision for BEV semantic mapping. A helicopter-mounted camera records high-resolution overhead RGB imagery (a), time-synchronized with an instrumented car that captures a forward-facing RGB view and LiDAR sweeps (b). By geo-aligning the aerial imagery to the vehicle frame, we obtain BEV-aligned crops and dense semantic targets that supervise BEV map prediction from ego sensors (c).}
  \label{fig:teaser}
  \vspace{-0.8em}
\end{figure*}

\section{Introduction}
\label{sec:intro}
Autonomous vehicles must reason about their surroundings in a representation that is both \emph{geometrically consistent} and \emph{actionable} for downstream tasks such as motion planning and map building \cite{stp3eccv22,fiery21}.
A common target is a \emph{scene-centric} bird’s-eye-view (BEV) raster—often called a \emph{dense BEV semantic environment map} or \emph{semantic occupancy grid}—that encodes drivable layout, static structure, and dynamic agents in a single top-down coordinate frame \cite{philion2020,openoccupancy23,occnet23}.
Compared to perspective-view perception, BEV maps align naturally with navigation, facilitate long-horizon reasoning, and are stable under viewpoint changes \cite{bevformer22,bevfusion23,beverse22}.

Despite major progress in BEV prediction from on-vehicle sensors (cameras and LiDAR) \cite{philion2020,bevformer22,bevfusion23}, \emph{supervision remains the key bottleneck} \cite{openoccupancy23,occ3d23,surroundocc23}.
Dense BEV labels are difficult to obtain from the ego platform alone: LiDAR is sparse and incomplete at long range; cameras observe the world in perspective and require non-trivial view transformation; and occlusions make “complete” BEV ground truth ambiguous without additional viewpoints \cite{occ3d23,surroundocc23}.
As a result, many BEV pipelines rely on proxy targets such as rasterized HD-map layers, projected 3D boxes, or temporally accumulated occupancy/semantic evidence \cite{hdmapnet22,maptr23,monolayout}.
These signals can be brittle for \emph{dynamic} objects and fine boundaries, and they often inherit biases from the ego sensor configuration and sensing range \cite{openoccupancy23,occ3d23}.
A different line of work pairs ground-level imagery with overhead imagery harvested from large-scale map/imagery sources, but these sources are typically \emph{not time-synchronized} with a given drive, leading to temporal inconsistencies that are especially problematic for learning dynamic scene elements \cite{torontocity17,workman15}.

In this paper, we revisit dense BEV supervision through a simple premise: \emph{overhead observation is an excellent teacher}, but only if it is temporally aligned with the driving scene.
We introduce \textbf{GOLD-BEV}, a framework and dataset that use \emph{time-synchronized} aerial imagery as training-time supervision for learning dense BEV semantic maps—including dynamic objects—from ego-centric sensors. 
Concretely, we capture simultaneous helicopter nadir RGB imagery and on-vehicle sensing (camera, multiple LiDARs, GNSS/INS), and convert the overhead view into BEV-aligned crops centered on the ego vehicle.
This aerial viewpoint provides (i) dense and topologically coherent context for static layout, and (ii) direct visibility of many dynamic agents that are hard to infer or label from ego sensors alone—\emph{without} requiring any overhead data at deployment time.

Aerial supervision also changes the human-annotation problem.
Rather than asking annotators to infer dense top-down semantics from sparse LiDAR or perspective images, we present \emph{aerial-style BEV imagery} directly in the target raster space, making semantic labeling intuitive and fast.
When time-synchronized aerial observations are available, we exploit them as dense supervision by producing BEV pseudo-labels with domain-adapted aerial teachers.
Crucially, we also train our BEV model to \emph{reconstruct} an aerial-style BEV RGB view from on-vehicle sensors, yielding an interpretable interface for rapid manual correction and systematic validation.
This reconstruction further enables scaling beyond aerial coverage: the model can generate \emph{pseudo-aerial} BEV images on unlabeled drives, which we use for low-budget annotation and uncertainty-aware pseudo-labeling, allowing the aerial teacher to bootstrap learning and then propagate supervision more broadly.

\paragraph{Contributions.}
\begin{enumerate}
    \item We propose a framework for directly inferring \emph{dense BEV semantic environment maps}, including dynamic objects, from on-vehicle sensors, and show that aerial-style BEV imagery enables intuitive semantic annotation with minimal manual effort.
    \item We leverage \emph{time-synchronized} aerial supervision, enabling accurate learning of dynamic scene elements and avoiding the temporal inconsistencies present in overhead sources that are not aligned with the drive. 
    \item We demonstrate scaling beyond aerial coverage by generating \emph{pseudo-aerial} BEV images, which support both human annotation and uncertainty-aware pseudo-labeling to train on unlabeled data. 
\end{enumerate}

\section{Related Work}
\label{sec:related}

\textbf{BEV semantic prediction from on-vehicle sensors.}
Bird's-eye-view (BEV) semantic mapping from surround cameras and/or LiDAR has become a standard representation for autonomous driving.
Lift-Splat-Shoot (LSS)~\cite{philion2020} pioneered end-to-end \emph{view transformation} by lifting image features into 3D and splatting them onto a rasterized BEV grid.
Subsequent work improved depth reliability and efficiency (e.g., BEVDepth~\cite{bevdepth22}) and introduced stronger spatiotemporal aggregation for camera-only BEV representations (e.g., BEVFormer~\cite{bevformer22}).
Multi-modal approaches unify features directly in BEV and support multiple perception tasks efficiently, e.g.\ BEVFusion~\cite{bevfusion23}.
Beyond dense raster maps, several methods target structured BEV outputs such as online vectorized HD-map elements; representative examples include MapTR~\cite{maptr23} and temporal/streaming variants such as StreamMapNet~\cite{streammapnet24}.

\textbf{Supervision and ground truth for BEV tasks.}
A persistent bottleneck for supervised BEV learning is high-quality ground truth.
Many datasets provide proxy BEV supervision from \emph{ego-centric} sources such as rasterized HD-map layers and 3D annotations (e.g., vehicles from 3D boxes), as commonly used in BEV segmentation and mapping pipelines~\cite{philion2020,hdmapnet22}.
Other works generate BEV/occupancy supervision through temporal sensor fusion and accumulation, e.g.\ MonoLayout~\cite{monolayout}, or build denser 3D/BEV labels via LiDAR-based pipelines with additional cleaning/annotation effort, e.g.\ OpenOccupancy~\cite{openoccupancy23}.
While effective, these signals are tied to the ego platform's sensing range, viewpoint, and/or availability of maps and annotations, and can be particularly brittle for dynamic agents and long-range context.

\textbf{Aerial viewpoints and cross-view learning.}
Aerial imagery has been widely used for semantic understanding and scene parsing, supported by datasets such as SkyScapes~\cite{Azimi_2019_ICCV} and UAVid~\cite{UAVid2020}.
Recent work has also explored the use of synthetic data, e.g., generated from simulation environments, as a complementary source for aerial image segmentation~\cite{synthetic_aerial_seg}.
Related cross-view literature often studies \emph{ground-to-aerial} matching and geo-localization (e.g., CVM-Net~\cite{Hu_2018_CVPR} and orientation-aware variants~\cite{Liu_2019_CVPR}), which differs from learning dense BEV semantics for driving.
More recently, aerial--ground cooperative driving datasets (e.g., AGC-Drive~\cite{agcdrive25}) provide multi-agent sensing and annotations, but typically target cooperative 3D perception (e.g., 3D boxes) rather than using aerial imagery as a \emph{training-time supervision signal} for dense BEV semantic maps.
In this work, we focus on time-synchronized aerial observation as a scalable source of dense BEV labels.

\section{Dataset}
\label{sec:dataset}
\begin{figure}[t]
\centering

\begin{subfigure}{0.2\linewidth}
    \centering
    \includegraphics[width=\linewidth]{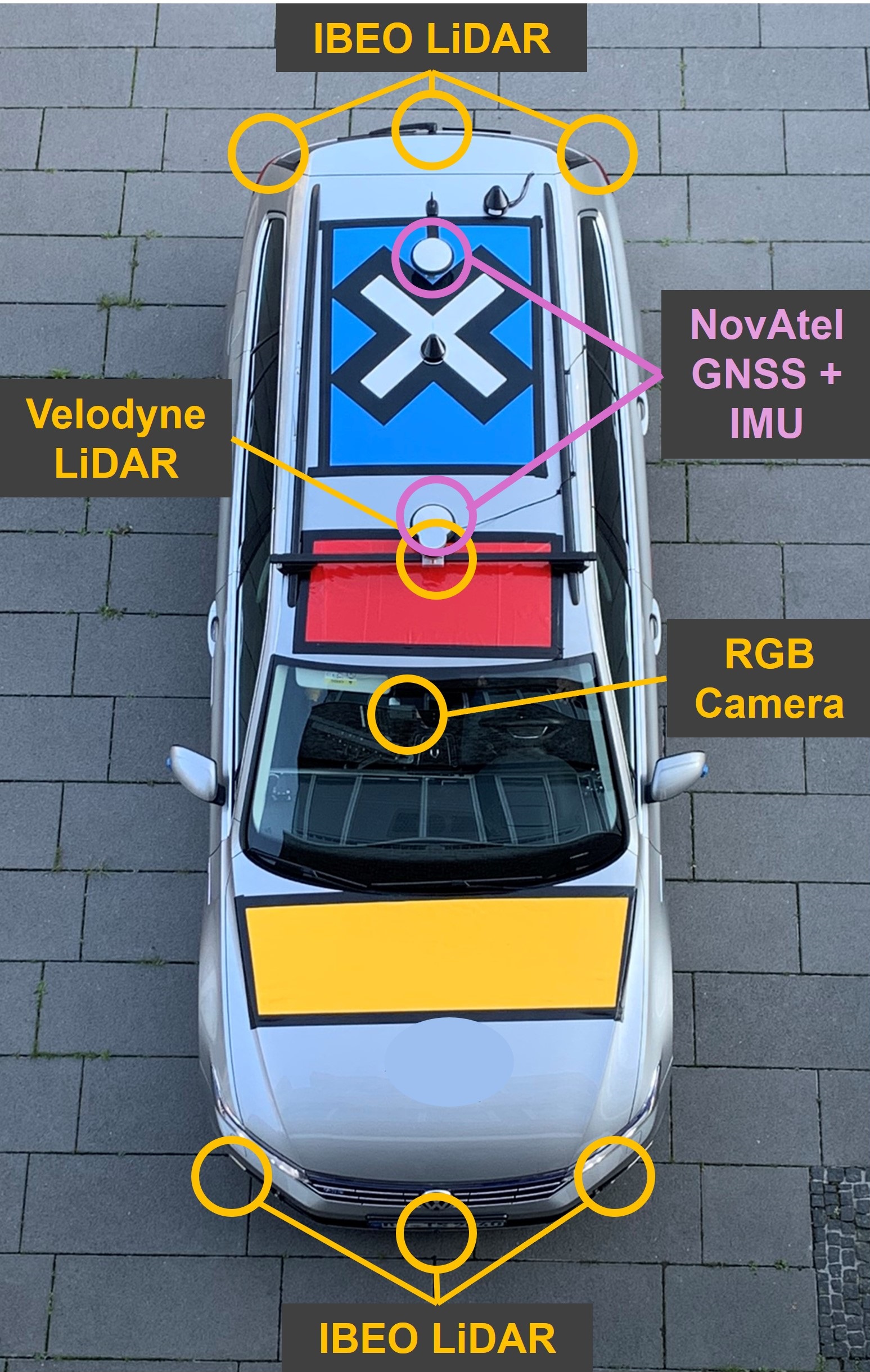}
    \caption{}
\end{subfigure}
\hfill
\begin{subfigure}{0.38\linewidth}
    \centering
    \includegraphics[width=\linewidth]{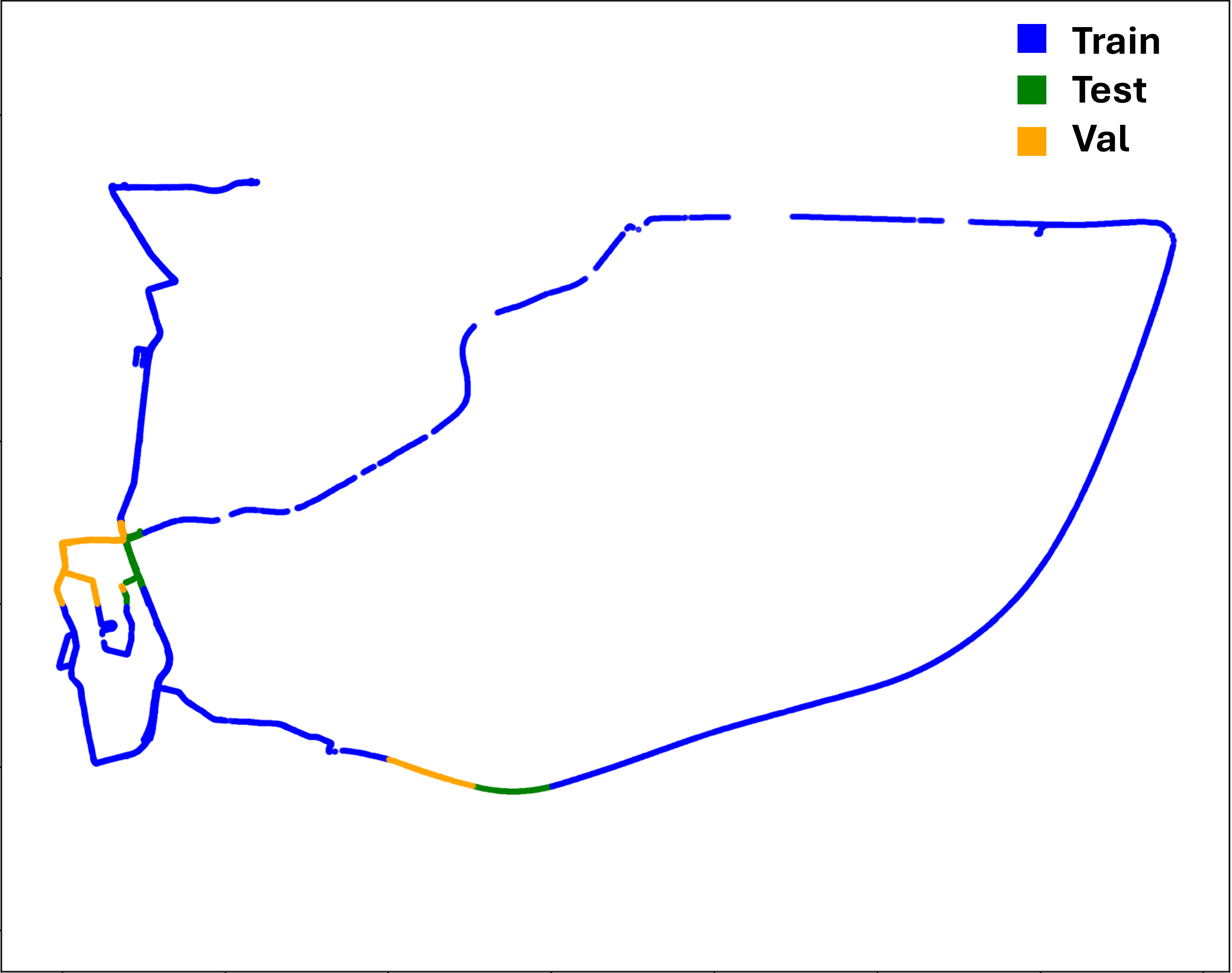}
    \caption{}
\end{subfigure}
\hfill
\begin{subfigure}{0.4\linewidth}
    \centering
    \includegraphics[width=\linewidth]{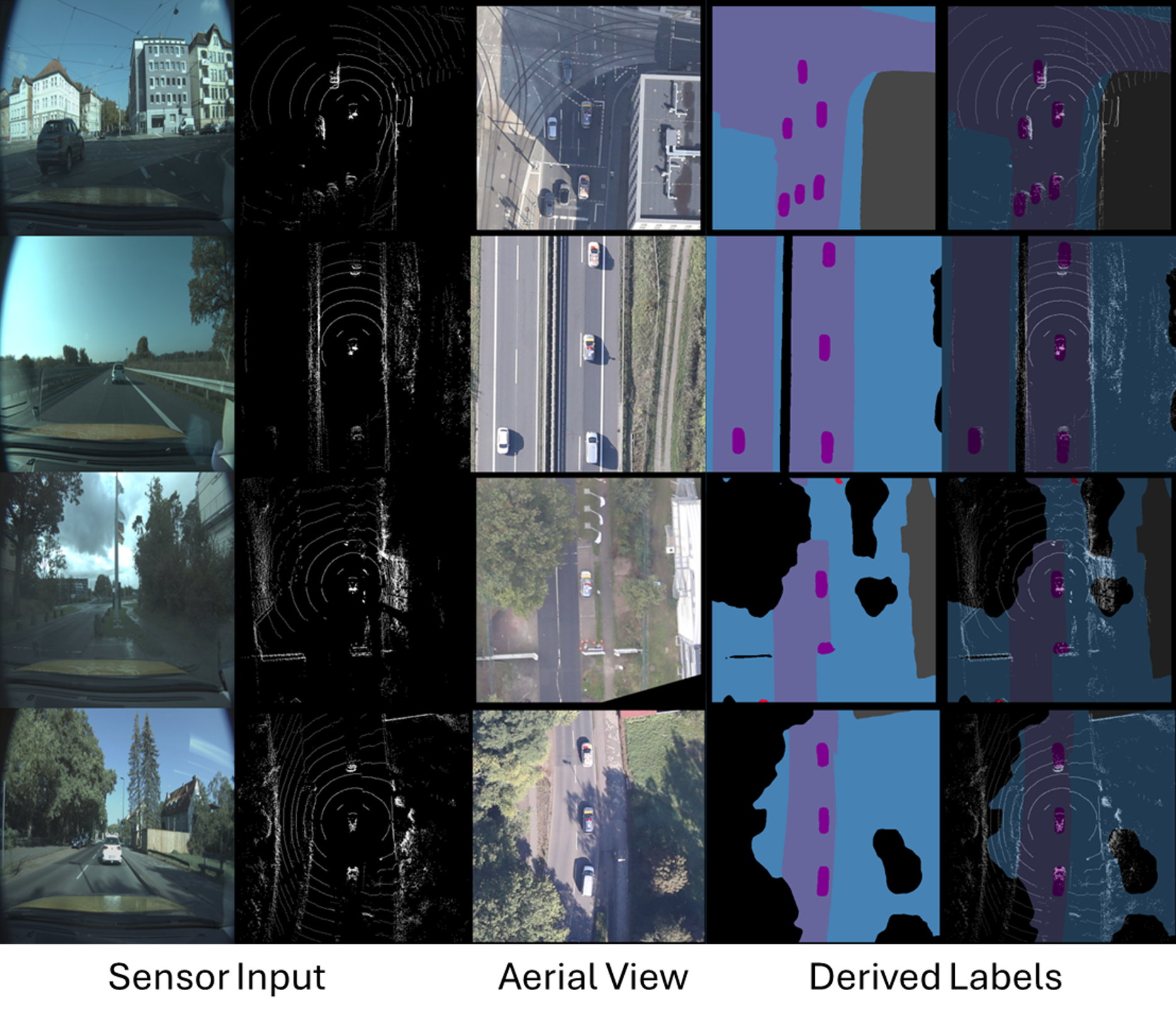}
    \caption{}
\end{subfigure}
\vspace{-0.3em}
  \caption{
  Vehicle setup used for the campaign and sensor installations (a). Trajectory of the data and how it is split into a train val test split (b).
  Examples for the input (camera and LiDAR), the corresponding aerial crops with the centered ego vehicle and the derived labels from the aerial view (c).
  }
  \label{fig:campaign_overview}
  \vspace{-0.8em}
\end{figure}

\paragraph{Overview.}
We introduce a large-scale cross-view dataset recorded simultaneously from an airborne helicopter platform and an instrumented measuring car in 
Germany, over three days in September 2022.
The vehicle covers \SI{60}{\kilo\meter} over roughly \SI{5}{\hour} across urban, suburban, and highway routes with varying traffic density.
After synchronization and quality filtering, the dataset contains \num{8199} samples, each comprising:
(i) a high-resolution aerial RGB image,
(ii) a forward-facing vehicle RGB image,
(iii) LiDAR sweeps from two vehicle-mounted sensors,
(iv) vehicle odometry, and
(v) GNSS/INS measurements.
See \autoref{fig:campaign_overview} for the platform setup and sensor placement.

\paragraph{Sensors.}
On the vehicle, we record with a forward-facing AVT Mako RGB camera, a Velodyne VLP-32C LiDAR, multiple IBEO automotive LiDAR sensors, and GNSS/IMU.
Aerial imagery is captured by a DSLR camera system mounted on the helicopter, producing RGB images of $3648\times5472$ pixels at $\sim$1\,Hz from an altitude of $\sim\SI{650}{\meter}$ (average GSD $\sim\SI{7}{\centi\meter}$/px); the aerial platform also logs GNSS/IMU.
The aerial and vehicle systems use independent time bases.

\subsection{Cross-view Alignment and BEV Cropping}
\label{sec:alignment_and_cropping}
We construct temporally and spatially aligned aerial--ground samples and derive vehicle-centric BEV-aligned aerial crops in a single preprocessing pipeline.

\paragraph{Temporal alignment.}
Each aerial image serves as a temporal anchor.
For every aerial frame, we select the closest vehicle RGB image with a maximum offset of 100\,ms, and then retrieve the closest Velodyne and IBEO LiDAR sweeps as well as GNSS/IMU measurements within 100\,ms of the selected vehicle image.
If any modality exceeds this threshold, the sample is discarded.
For retained samples, mean offsets are 12\,ms (vehicle RGB), 23\,ms (Velodyne), 10\,ms (IBEO), and 5\,ms (GNSS/IMU), limiting motion-induced misalignment.

\paragraph{Spatial alignment.}
Given the temporally aligned sensor bundle, we localize the ego vehicle in each aerial image.
We first project the vehicle position into the aerial image using GNSS from both platforms.
To refine this estimate---particularly in GNSS-challenging urban areas---we apply template matching in a window around the projected location; the vehicle roof was marked with an ``X'' pattern to facilitate detection.
We retain detections with confidence above 0.65 and discard all remaining samples.

\paragraph{BEV-aligned aerial crops.}
To obtain vehicle-centric overhead context in a consistent raster space, we rotate each aerial image so that the vehicle heading points upward, using IMU data from both platforms.
We then crop a $600\times600$ patch centered on the vehicle, corresponding to roughly $42\times42$\,m at the average GSD.
These BEV-aligned crops define the coordinate frame used in the paper.

\subsection{Supervision Signals}
\label{sec:dataset_supervision}
The BEV-aligned aerial crops provide (i) a dense overhead RGB target in BEV space and (ii) training-time supervision for BEV semantic mapping.
We generate dense BEV pseudo-labels with domain-adapted aerial teachers and conservative fusion; details are given in \autoref{sec:methods_uda_teachers_fusion}.  \autoref{fig:campaign_overview} (c) shows examples of such pseudo labels.
Additionally, we provide sparse semantic supervision by manually annotating \num{454} Velodyne LiDAR sweeps (used for the LiDAR holdout evaluation and optional refinement; see \autoref{sec:methods_training}).

\paragraph{Semantic taxonomy.}
We use five classes: \textit{road}, \textit{sidewalk}, \textit{building}, \textit{vehicle}, and \textit{VRU}.
\textit{Road} covers drivable roadway surfaces.
\textit{Sidewalk} denotes other flat, traversable ground surfaces that are not part of the roadway, such as sidewalks, paved shoulders, plazas, parking areas, low vegetation and bike/foot paths.
\textit{Building} covers buildings and other permanent vertical structures.
\textit{Vehicle} includes cars, trucks, buses, and motorcycles.
\textit{VRU} (vulnerable road users) includes pedestrians, e-scooter, motorcycle, wagon and cyclists.

\subsection{Dataset Splits}
\label{sec:splits}
To prevent spatial leakage, we split data by vehicle trajectory segments (geographic partitioning) rather than random sampling, cf.\ \autoref{fig:campaign_overview}(b).
This yields \textcolor{black}{6510} train, \textcolor{black}{792} validation, and \textcolor{black}{897} test samples, each containing a mix of urban and highway scenes.

\subsection{Comparison to Existing Cross-View Datasets}
\label{sec:dataset_comparison}
\autoref{tab:data_comparison} summarizes key differences to prior cross-view benchmarks.
In contrast to web-sourced street--satellite pairs and vehicle logs paired with non-synchronized overhead imagery, our data is captured simultaneously from aerial and ground platforms and includes multi-sensor vehicle measurements, enabling temporally consistent supervision of dynamic traffic participants.

\begin{table}[t]
\centering
\caption{Comparison with existing datasets. Multi-sensor includes LiDAR/INS.}
\label{tab:data_comparison}

\renewcommand{\arraystretch}{1.2}
\setlength{\tabcolsep}{4pt}

\resizebox{\linewidth}{!}{%
\begin{tabular}{lcccc}
\toprule
Dataset & Overhead Source & Simult. Aerial--Ground & Multi-Sensor & Metric BEV Alignment \\
\midrule
CVUSA\cite{workman2015localize}/CVACT\cite{Liu_2019_CVPR} & Web satellite maps & \xmark & \xmark & \xmark \\
KITTI-CVL\cite{wang2023satellite} & Google Maps & \xmark & \cmark\ KITTI\cite{GeigerLenzStiller2013_1000040205} & \cmark \\
FordAV-CVL\cite{wang2023satellite} & Google Maps & \xmark & \cmark\ FordAV\cite{Agarwal_2020} logs & \cmark \\
nuScenes\cite{nuscenes2019} & None & \xmark & \cmark & N/A \\
\textbf{Ours} & Helicopter aerial capture & \textbf{\cmark} & \textbf{\cmark} & \textbf{\cmark} \\
\bottomrule
\end{tabular}
}

\end{table}

\section{Methods}
\label{sec:method}

Given time-synchronized on-vehicle sensors (front-view RGB camera and/or LiDAR) and a temporally aligned aerial observation during training,
our goal is to predict a \emph{dense} semantic bird's-eye-view (BEV) map centered at the ego vehicle.
We formulate BEV segmentation as per-cell semantic classification on a fixed raster grid aligned with the vehicle heading.
In addition to BEV semantics, we optionally predict a \emph{pseudo-aerial} BEV RGB image in the same raster space, enabling multi-task learning
and providing an interpretable interface for lightweight manual annotation.
The following sections will describe the architectures, creation of aerial pseudo labels and training strategies that were used.

\subsection{Architectures for BEV Segmentation and Reconstruction}
\label{sec:methods_arch}
We created two architecutes for BEV segmentation and reconstruction that receive the same paired inputs: a forward-facing RGB image and a LiDAR point cloud rendered into a BEV raster aligned to the target BEV grid.

\paragraph{LiDAR-to-BEV rasterization.}
We convert each LiDAR sweep into a fixed-size bird's-eye-view tensor by discretizing the ground plane into a regular BEV grid (identical resolution and spatial extent as the semantic output).
All points are transformed into the ego frame, cropped to the BEV region of interest, and accumulated per cell.
For each BEV cell we compute three channels: (i) \emph{occupancy} (binary indicator of at least one return),
(ii) \emph{height} (e.g., maximum $z$ value within the cell, normalized to a fixed range), and
(iii) \emph{density} (log-scaled point count per cell).
The resulting 3-channel tensor (similar to e.g. \cite{chen2017mv3d,ku2017avod})
serves as the LiDAR input to all models.

\paragraph{Camera preprocessing.}
The RGB image is resized to the network input resolution of $600\times 600$, and encoded by a dedicated camera backbone.

\paragraph{The SegFormer architecture:}
Our BEV student uses two SegFormer encoders (camera and LiDAR-BEV) that output four stage feature maps each.
The LiDAR stream ingests a 3-channel BEV raster (occupancy, height, log-density), while the camera stream uses the forward RGB image. 
To fuse modalities, we apply cross-attention fusion blocks at the deeper stages (S3/S4): BEV features act as queries, while camera features provide keys/values (similar to~\cite{guan2026crfusion,bai2022transfusion}). The attention output is projected back to the BEV channel dimension and combined with the input BEV features using a residual connection followed by normalization, as in standard transformer blocks~\cite{vaswani2017attention}.
A SegFormer-style lightweight decoder head projects each stage to a shared hidden dimension, upsamples to a common resolution, concatenates, and fuses them with a $1{\times}1$ convolution to predict dense BEV logits at full resolution. 
Optionally, we attach (i) a shallow BEV reconstruction head on the fused decoder feature to output a deterministic pseudo-aerial BEV RGB image.

\paragraph{Diffusion reconstructor.}
For higher-fidelity pseudo-aerial BEV imagery, we train a conditional latent diffusion model based on Stable Diffusion v1.5 with ControlNet.
We freeze the pretrained VAE and UNet~\cite{rombach2022ldm} and train (i) a ControlNet branch~\cite{zhang2023controlnet}
and (ii) a lightweight sensor-to-conditioning network.
The conditioner mirrors our two-stream SegFormer-XCA encoder/fusion but outputs a compact $C$-channel conditioning map (typically $C{=}3$) that is injected into the frozen UNet via ControlNet residuals.
Training follows the standard latent noise-prediction objective on VAE latents, using a mask to restrict supervision to valid BEV cells.
At inference, we sample with a DDIM scheduler for a fixed number of steps and decode the final latent with the frozen VAE.
\emph{Auxiliary segmentation head.}
To encourage semantic consistency, we optionally attach a lightweight segmentation head to the final feature map of the (frozen) VAE decoder.
The head consists of two $3{\times}3$ Conv-GN-SiLU blocks followed by a $1{\times}1$ convolution that maps features to $K$ semantic classes.
During training, we decode the current denoised latent estimate (using the same predicted-clean latent used for reconstruction) through the VAE decoder,
apply the segmentation head to its last feature map, and supervise with a per-pixel cross-entropy (and/or Dice) loss on valid BEV cells.
The segmentation loss is optimized jointly with the diffusion noise-prediction loss; the auxiliary head is trained only during diffusion training and is not required at inference.

\subsection{Domain-Adapted Aerial Teachers and Pseudo-Label Fusion}
\label{sec:methods_uda_teachers_fusion}

\paragraph{Overview.}
We obtain dense BEV supervision by training two \emph{aerial} semantic segmentation teachers on BEV-aligned aerial crops and fusing their
predictions into a single pseudo-label map per crop.
To reduce the appearance gap between public UAV imagery and our helicopter data, we adapt both teachers to our target domain
using semi-supervised domain adaptation on a mixture of labeled and unlabeled target crops.

\paragraph{Teacher supervision sources.}
We create the teachers from two supervised signals:
\begin{enumerate}
    \item \textbf{Labeled target subset (train/val).}
    We manually annotate a subset of our BEV-aligned aerial crops and split it into a small \emph{training} and \emph{validation} set.
    The training split is used for supervised fine-tuning of both teachers on the target domain, while the validation split is used for model selection,
    calibration (label mapping and fusion thresholds), and early stopping.
    The target labels cover (i) a coarse structural taxonomy (road/sidewalk/building/vehicle) and (ii) a pedestrian-specific mask.
    \item \textbf{External supervised pre-training.}
    We pre-train the structural teacher using the SkyScapes aerial dataset (after mapping its label space to our taxonomy),
    and pre-train the pedestrian teacher on a proprietary pedestrian segmentation dataset.
\end{enumerate}
Both teachers employ a Mask2Former-style~\cite{cheng2021mask2formervideoinstancesegmentation}  segmentation backbone.

\paragraph{Semi-supervised domain adaptation to helicopter crops.}
Starting from the external supervised initialization, we adapt each teacher to our target helicopter crops using a semi-supervised recipe:
we combine (i) standard supervised training on the labeled target \emph{training} split with (ii) SSL/UDA losses on unlabeled target crops
(e.g., consistency regularization under strong augmentation and self-training on high-confidence pseudo-labels).
We select checkpoints and tune thresholds using the labeled target \emph{validation} split, following common practice in semi-supervised segmentation and unsupervised domain adaptation \cite{Schwonberg2023Survey,niemeijer2021combining}.
The resulting adapted checkpoints are then applied to all aerial crops to generate dense pseudo-labels for. 

\paragraph{Producing teacher predictions.}
Given a BEV crop image $\mathbf{x}$ (600$\times$600), the structural teacher outputs per-pixel class probabilities
$p_c^k(\mathbf{x})$ over $K{=}4$ classes. We extract the argmax label and confidence as
\begin{equation}
\hat{y}_c(\mathbf{x}) = \arg\max_k p_c^k(\mathbf{x}), \qquad
\hat{s}_c(\mathbf{x}) = \max_k p_c^k(\mathbf{x}).
\end{equation}
The pedestrian teacher is trained as a binary model and outputs a pedestrian probability $p_{\text{ped}}(\mathbf{x})\in[0,1]$.

\paragraph{Fusion into final pseudo-labels.}
We fuse both teachers with a conservative tri-state policy that prioritizes reliable dynamic labels and ignores ambiguous pixels.
Let $\tau_{\text{ped}}^{\text{hi}}$ and $\tau_{\text{ped}}^{\text{lo}}$ be pedestrian high/low thresholds and $\tau_c$ a structural confidence threshold.
The fused label map $\hat{y}(\mathbf{x})$ is defined per pixel as
\begin{equation}
\hat{y}(\mathbf{x}) =
\begin{cases}
\text{PED}, & p_{\text{ped}}(\mathbf{x}) \ge \tau_{\text{ped}}^{\text{hi}}, \\[2pt]
\hat{y}_c(\mathbf{x}), & p_{\text{ped}}(\mathbf{x}) \le \tau_{\text{ped}}^{\text{lo}} \ \wedge\ \hat{s}_c(\mathbf{x}) \ge \tau_c, \\[2pt]
\text{IGNORE}, & \text{otherwise.}
\end{cases}
\end{equation}
Thus, confident pedestrian predictions overwrite the structural map, while structural labels are accepted only above $\tau_c$; all ambiguous regions are ignored.

\paragraph{Ignoring trees.}
Pixels predicted as tree/vegetation are always mapped to \texttt{IGNORE} and excluded from training.
In our helicopter imagery, tree canopies often occlude or overlay regions that are nevertheless visible and semantically relevant from the ground,
making aerial tree labels a frequent source of incorrect supervision.

\subsection{Training with Dense Pseudo-Labels and Sparse LiDAR Supervision}
\label{sec:methods_training}

\paragraph{Overview.}
Given BEV-aligned aerial crops paired with dense BEV pseudo-labels and a corresponding BEV RGB target, we train the architectures from Sec.~\ref{sec:methods_arch} to jointly predict (i) dense BEV semantics and (ii) a pseudo-aerial BEV reconstruction.
Training proceeds in two stages: (1) dense supervision from aerial-derived pseudo-labels and (2) an refinement stage using LiDAR semantic annotations.

\paragraph{Stage 1: Dense pseudo-label training (aerial supervision).}
For BEV semantic segmentation, we supervise the BEV logits with a per-pixel cross-entropy loss on all non-ignore cells of the pseudo-label map.
For deterministic reconstruction (SegFormer decoder), we apply an image-level $\ell_1$ loss between the predicted BEV RGB and the BEV RGB target.
For diffusion-based reconstruction, we train a ControlNet-conditioned denoiser with the standard noise-prediction objective in latent space.

\paragraph{Stage 2 (optional): Sparse LiDAR refinement.}
Dense aerial pseudo-labels can be noisy, in particular for dynamic agents and at fine boundaries.
To inject geometry-grounded semantics and correct residual misalignments, we optionally fine-tune the segmentation head using sparse LiDAR semantic annotations.
We rasterize the labeled LiDAR points to the BEV grid and compute the segmentation loss only on supervised BEV cells.
This stage serves two purposes: (i) it improves dynamic-class accuracy where aerial pseudo-labels are uncertain, and (ii) it mitigates small systematic offsets between projected LiDAR measurements and the BEV raster/imagery.

\section{Experiments}
\label{sec:experiments}

\begin{table}[t]
\centering
\caption{Main results on the Gold BEV Test split (dense BEV segmentation). \textbf{L} indicates an additional fine-tuning stage using sparse LiDAR semantic annotations. LiDAR Holdout mIoU is computed on supervised cells only. For camera-only models, all metrics are computed inside the BEV visibility cone (pixels outside are ignored). 
}
\label{tab:main_results}
\scriptsize
\setlength{\tabcolsep}{2pt}
\renewcommand{\arraystretch}{1.1}

\resizebox{\columnwidth}{!}{%
\begin{tabular}{L{2.45cm} C{0.8cm} C{0.55cm}
                C{0.8cm} C{0.8cm} C{0.8cm} C{0.8cm}
                C{0.85cm} C{0.85cm} C{0.85cm} C{0.85cm}}
\toprule
Method & Inp. & \hdr{L} &
\multicolumn{4}{c}{\textbf{Gold BEV Test (seg.)}} &
\multicolumn{4}{c}{\textbf{LiDAR Holdout (seg.)}} \\
\cmidrule(lr){4-7}\cmidrule(lr){8-11}
& & &
\hdr{mIoU\\All} & \hdr{mIoU\\ Static} & \hdr{IoU\\Vehic.} & \hdr{IoU\\VRU.} &
\hdr{mIoU\\All} & \hdr{mIoU\\ Static} & \hdr{IoU\\Vehic.} & \hdr{IoU\\VRU.} \\
\midrule

SegF.      & L   & --         & 0.485 & 0.627 & 0.506 & 0.039 & 0.426 & 0.549 & 0.427 & 0.056 \\
SegF.      & C   & --         &0.290 & 0.408 & 0.226 & 0.000 & 0.168 & 0.280 & 0.003 & 0.000 \\
\midrule

SegF.       & C+L & $\times$    &0.477 & 0.616 & 0.500 & 0.035 & 0.442 & 0.559 & 0.452 & 0.084 \\
SegF. (Int.)& C+L & \checkmark  &0.461 & 0.591 & 0.494 & 0.036 & 0.656 & 0.709 & 0.854 & 0.299 \\
SegF. (EP.) & C+L & \checkmark  &0.461 & 0.617 & 0.415 & 0.039 & 0.690 & 0.746 & 0.880 & 0.328 \\
\midrule

Diff. & C+L & $\times$ & 0.485 & 0.663 & 0.438 & 0.0003 & 0.437 & 0.593 & 0.404 & 0.0004 \\
\bottomrule
\end{tabular}
}
 \vspace{-0.8em}
\end{table}

We evaluate along the paper’s three claims: (i) dense BEV semantic mapping from ego sensors, (ii) the impact of \emph{time-synchronized} aerial supervision on dynamic classes, and (iii) scaling beyond aerial coverage via pseudo-aerial BEV imagery for annotation and pseudo-labeling.
Unless stated otherwise, datasets, label generation, and model architectures follow Sections~\ref{sec:dataset} and~\ref{sec:method}.

\vspace{0.25em}
\noindent\textit{Splits.}
We report final numbers on a held-out \emph{Gold} test split; all model selection uses disjoint validation data.

\vspace{0.25em}
\noindent\textit{Sparse LiDAR holdout.}
We additionally evaluate on a LiDAR-annotated holdout split (10\% of the data c.f. Sec.~\ref{sec:dataset}) by rasterizing point-wise labels to BEV and computing mIoU on supervised cells only (c.f. Sec.~\ref{sec:method}). 

\vspace{0.25em}
\noindent\textit{Metrics.}
BEV segmentation is measured by mIoU over all classes (mIoU\textsubscript{All}) and static classes (road, sidewalk, building mIoU\textsubscript{Static}); dynamic performance is reported via per-class IoU for Vehicle and VRU (and mIoU\textsubscript{Dyn.} where applicable).
Reconstruction quality is reported using PSNR/SSIM, LPIPS, and FID/KID against the BEV reference imagery.

\vspace{0.25em}
\noindent\textit{Low-budget setting.}
For the no-aerial scenario we assume a 10-frame manual BEV annotation budget on pseudo-aerial frames (Sec.~\ref{sec:without_aerial_supervision}).

\subsection{Results on Our Dataset}
\label{sec:results_ours}

Our dataset supports supervised learning of two complementary BEV tasks from on-board sensors (camera and/or LiDAR):
(i) reconstructing a pseudo BEV image and (ii) predicting dense BEV semantic segmentation.
Table~\ref{tab:main_results} reports segmentation results on the Gold BEV test split and on the sparse LiDAR holdout.

\paragraph{Training setup and ablations.}
All models are trained in a multi-task setting to jointly predict BEV reconstruction and BEV semantic segmentation.
We study three input configurations: camera-only (C), LiDAR-only (L), and camera+LiDAR (C+L).
For camera-only models, supervision and evaluation are restricted to the camera-visible BEV region:
we apply a forward BEV visibility-cone mask during training and evaluation, and ignore pixels outside the cone for both losses and metrics.
For the sparse LiDAR holdout, point-wise labels are rasterized into the BEV grid using the calibrated projection described in Sec.~\ref{sec:methods_training}, and mIoU is computed on supervised cells only.

\paragraph{Effect of input modality.}
LiDAR-only provides a strong estimate of static layout on Gold (mIoU\textsubscript{Static}$=0.627$) and achieves solid overall segmentation (mIoU\textsubscript{All}$=0.485$), with high Vehicle IoU ($0.506$).
Camera-only performs markedly worse (mIoU\textsubscript{All}$=0.290$ within the visible cone), indicating that monocular appearance cues are insufficient to recover BEV semantics under partial observability.
Fusing camera and LiDAR improves balance across splits: SegFormer C+L achieves comparable Gold performance (mIoU\textsubscript{All}$=0.477$) while improving LiDAR-holdout mIoU\textsubscript{All} over single-modality baselines ($0.442$ vs.\ $0.426$ for L-only). 

\paragraph{SegFormer vs.\ diffusion for BEV semantics.}
The diffusion-based multi-task model attains competitive overall segmentation on Gold (mIoU\textsubscript{All}$=0.485$) and yields the strongest static score among the compared models (mIoU\textsubscript{Static}$=0.663$).
However, its dynamic-class performance is weaker (Vehicle IoU $0.438$) and VRU IoU is near-zero on both Gold and the LiDAR holdout.
In contrast, SegFormer variants maintain more reliable dynamic predictions (e.g., SegFormer C+L Vehicle IoU $0.500$ and LiDAR-holdout VRU IoU $0.084$), suggesting that diffusion excels at learning scene layout and appearance but is less effective as a semantic decoder for fine-scale dynamic agents without geometry-grounded supervision.

\begin{figure}[t]
  \centering
  \includegraphics[width=\linewidth]{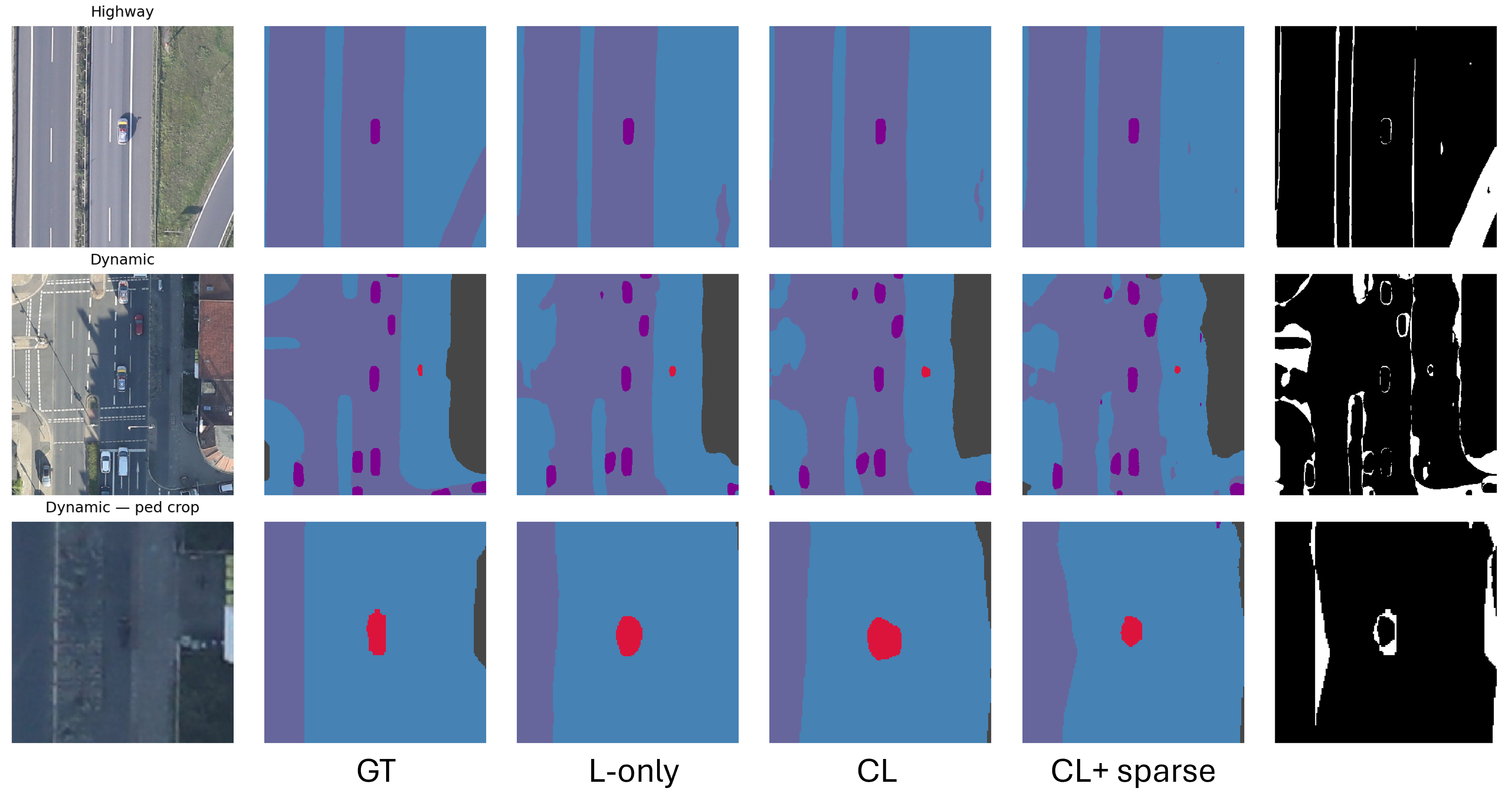}
  
  \caption{
  Comparison of BEV semantic segmentation predictions.
  Each row shows a different scenario: a highway scene (top), a complex urban intersection (middle), and a zoomed-in crop highlighting a VRU prediction (bottom).
  From left to right: ground truth (GT), LiDAR-only prediction (L-only), camera+LiDAR prediction (C+L), and camera+LiDAR with additional sparse LiDAR fine-tuning (C+L + sparse).
  }
  \label{fig:bev_seg_qual}
  \vspace{-0.8em}
\end{figure}

\paragraph{Sparse LiDAR supervision improves geometry-consistent semantics.} Adding sparse LiDAR supervision yields large gains on the LiDAR holdout split, especially for dynamic agents.
Importantly, the two evaluations differ: Gold BEV metrics assess \emph{dense} predictions over the full raster (including regions that may be occluded or outside LiDAR visibility), whereas the LiDAR holdout measures agreement only on \emph{actually observed} cells with rasterized point-wise labels.
Compared to SegFormer C+L without sparse supervision (LiDAR holdout mIoU\textsubscript{All}$=0.442$), both refinement strategies substantially improve visible-region performance:
interleaved sparse supervision reaches $0.656$ and epoch-level fine-tuning reaches $0.690$.
The gains are driven by dynamic classes (Vehicle IoU $0.452 \rightarrow 0.854/0.880$; VRU IoU $0.084 \rightarrow 0.299/0.328$),
showing that a small amount of geometry-grounded supervision quickly injects reliable dynamic semantics that are difficult to learn from aerial-derived dense targets alone.
This is particularly pronounced for VRUs: in aerial BEV crops they occupy only a few pixels, making dense pseudo-labels noisy and recall-limited, while sparse LiDAR supervision provides a clean visibility-filtered signal that can be incorporated effectively through a short fine-tuning stage.
On the Gold split, sparse supervision does not necessarily increase agreement with aerial-derived labels (mIoU\textsubscript{All} $0.477 \rightarrow 0.461$),
which is consistent with the supervision mismatch: sparse LiDAR optimizes correctness on visible geometry, while dense aerial targets also score predictions in unobserved regions where the label is inherently less certain.

\paragraph{Reconstruction quality.}
\autoref{tab:bev_recon_metrics} quantifies pseudo-aerial BEV reconstruction quality.
SegFormer achieves higher pixel-level fidelity (PSNR/SSIM: $14.72/0.509$ vs.\ $13.50/0.505$),
while diffusion substantially improves perceptual and distributional quality (LPIPS: $0.803 \rightarrow 0.596$, FID: $282.2 \rightarrow 140.1$, KID: $0.301 \rightarrow 0.073$).
Qualitatively (\autoref{fig:pseudo_aerial_views}), SegFormer reconstructions tend to be smoother, whereas diffusion produces sharper structures and more realistic local texture.
This motivates using pseudo-aerial imagery primarily as an annotation interface, where perceptual realism and crisp boundaries are beneficial, while keeping semantics grounded via explicit segmentation supervision.

\begin{table}[t]
\centering

\begin{minipage}[t]{0.49\linewidth}
\centering
\caption{\textbf{Reconstruction quality on BEV RGB (test split).} Pixel-level (PSNR/SSIM), perceptual (LPIPS), and distributional (FID/KID) metrics between reconstructed BEV and GT BEV. Best results are in bold.}
\label{tab:bev_recon_metrics}
\scriptsize
\setlength{\tabcolsep}{3pt}
\renewcommand{\arraystretch}{1.1}

\resizebox{\linewidth}{!}{%
\begin{tabular}{@{}L{1.1cm} C{1.1cm} C{1.1cm} C{1.1cm} C{0.85cm} C{0.85cm}@{}}
\toprule
Method &
\hdr{PSNR$\uparrow$} &
\hdr{SSIM$\uparrow$} &
\hdr{LPIPS$\downarrow$} &
\hdr{FID$\downarrow$} &
\hdr{KID$\downarrow$} \\
\midrule
SegF. &
\textbf{14.72 $\pm$ 1.28} &
\textbf{0.509 $\pm$ 0.050} &
0.803 $\pm$ 0.033 &
282.2 & 0.301 \\
Diff. &
13.50 $\pm$ 0.93 &
0.505 $\pm$ 0.048 &
\textbf{0.596} $\pm$ \textbf{0.023} &
\textbf{140.1} & \textbf{0.073} \\
\bottomrule
\end{tabular}%
}
\end{minipage}
\hfill
\begin{minipage}[t]{0.49\linewidth}
\centering
\caption{Sparse supervision efficiency. We vary the fraction of LiDAR-supervised training frames. C+L+sparse (epoch FT).}
\label{tab:abl_sup_fraction}
\scriptsize
\setlength{\tabcolsep}{2pt}
\renewcommand{\arraystretch}{1.1}

\resizebox{\linewidth}{!}{%
\begin{tabular}{@{}L{1.0cm} C{0.95cm} C{0.85cm} C{0.85cm} C{0.85cm} C{0.85cm}@{}}
\toprule
Variant & Sup frac. &
\hdr{Gold\\mIoU\\All} &
\hdr{Gold\\Dyn.} &
\hdr{LiDAR\\mIoU\\All} &
\hdr{LiDAR\\Dyn.} \\
\midrule
C+L  & 0.10 & \textbf{0.467} & \textbf{0.227} & 0.682 & 0.603  \\
C+L  & 0.25 & 0.457 & 0.225 & \textbf{0.696} & \textbf{0.628}  \\
C+L  & 0.50 & 0.463 & \textbf{0.227} & 0.670 & 0.585  \\
C+L  & 1.00 & 0.461 & \textbf{0.227} & 0.690 & 0.604   \\
\bottomrule
\end{tabular}%
}
\end{minipage}

\end{table}

\subsection{Influence of Supervised Labels}
\label{sec:influence_sup_labels}

We study how much \emph{sparse} LiDAR semantic supervision is needed to refine dense BEV predictions.
Concretely, we vary the fraction of LiDAR-supervised training frames used in the epoch-level fine-tuning stage (C+L+sparse, epoch FT) and report Gold performance and LiDAR-holdout mIoU computed on supervised cells only (\autoref{tab:abl_sup_fraction}).

\paragraph{Sparse supervision is label-efficient.}
A small supervision budget already achieves near-peak performance.
Using only 10\% of LiDAR-supervised frames yields the best Gold mIoU\textsubscript{All} ($0.467$) while reaching strong LiDAR-holdout performance ($0.682$).
At 25\%, LiDAR-holdout peaks (mIoU\textsubscript{All}$=0.696$, mIoU\textsubscript{Dyn.}$=0.628$), indicating that limited geometry-anchored supervision suffices to correct systematic failure modes.
\paragraph{Diminishing returns beyond 25\%.}
Gold performance remains essentially unchanged (mIoU\textsubscript{All} $0.457$--$0.463$; mIoU\textsubscript{Dyn.} $\approx 0.225$--$0.227$),
and LiDAR-holdout varies mildly with no monotonic trend.
Overall, sparse supervision primarily acts as a targeted refinement signal: it improves LiDAR-consistent semantic correctness without requiring dense manual BEV labels.

\begin{table}[t]
\caption{Intersection-over-Union (IoU) evaluation of individual annotators and their fusion compared to the ground truth. The column \emph{Vehicle vis.} reports the IoU computed only on vehicles visible in LiDAR, while \emph{Vehicle} includes all annotated vehicles. Bold values indicate the best performance per column.}
\label{tab:bev_iou_visible_vs_all}
\centering
\scriptsize
\setlength{\tabcolsep}{3pt}
\renewcommand{\arraystretch}{1.15}
\begin{tabular}{lccccccc}
\toprule
Annotator & mIoU & Ignored & Road & Sidewalk & Building & Vehicle vis. & Vehicle \\
\midrule
A1 & 0.747 & 0.276 & \textbf{0.919} & \textbf{0.879} & 0.648 & 0.541 & 0.560  \\
A2 & \textbf{0.749} & 0.337 & 0.874 & 0.801 & \textbf{0.800} & 0.523 & 0.520\\
A3 & 0.551 & 0.268 & 0.804 & 0.504 & 0.395 & 0.501 & 0.502  \\
A4 & 0.748 & 0.440 & 0.906 & 0.802 & 0.747 & \textbf{0.539} & \textbf{0.561}  \\
Fusion & \textbf{0.896} & 0.642 & \textbf{0.966} & \textbf{0.917} & \textbf{0.990} & \textbf{0.712} & \textbf{0.708}  \\

\bottomrule
\end{tabular}
 \vspace{-0.8em}
\end{table}

\subsection{What to do Without Aerial Supervision}
\label{sec:without_aerial_supervision}
Time-synchronized aerial supervision is not always available outside covered regions.
To scale beyond aerial coverage (Contribution~3), we generate \emph{pseudo-aerial} BEV imagery and use it
(i) to obtain a small amount of direct BEV supervision via lightweight manual labeling and
(ii) to exploit additional unlabeled target frames via uncertainty-aware pseudo-labeling.

\vspace{0.25em}
\noindent\textit{10-frame manual BEV supervision (two-view annotation + strict agreement fusion).}
We annotate 10 pseudo-aerial frames in the same BEV raster as our predictor.
Pseudo-aerial reconstructions are reliable for \emph{static layout} and \emph{vehicles}, but \emph{VRU are not practically annotatable}:
they occupy only a few pixels in overhead crops and are often indistinguishable from texture/shadows in the reconstructions.
We therefore omit VRUs in the manual pseudo-aerial masks and inject VRU semantics via LiDAR-based signals (below).

\begin{figure}[t]
  \centering
  \includegraphics[width=\linewidth]{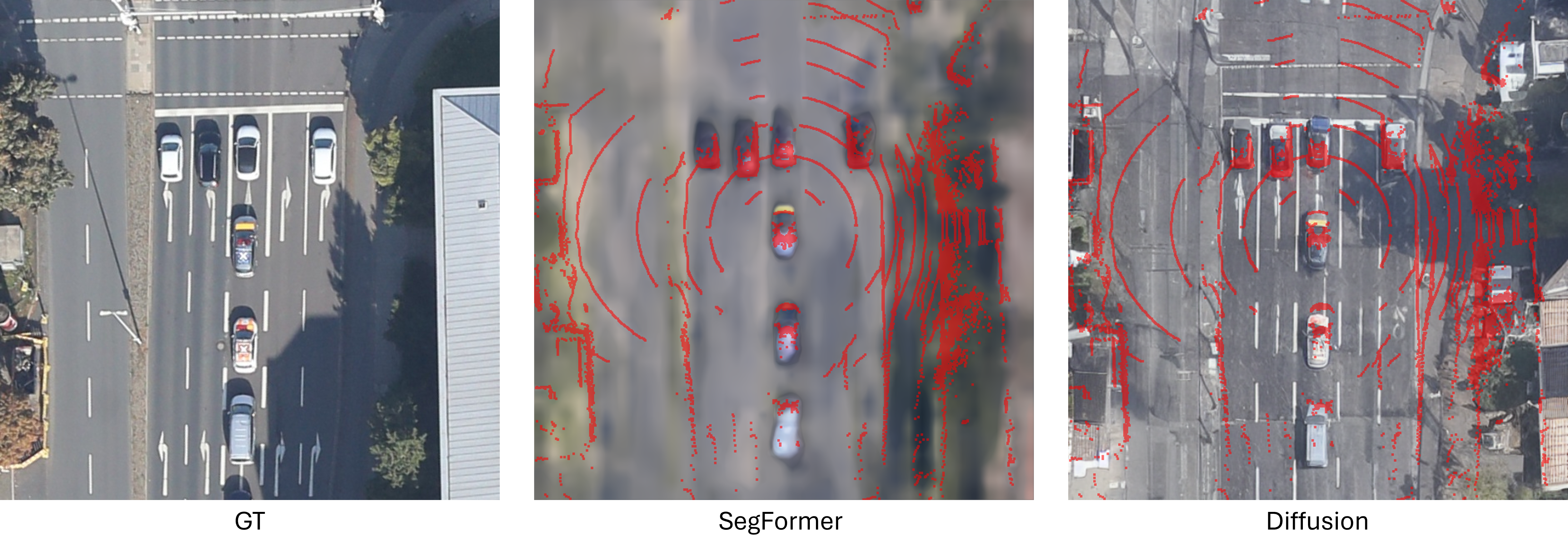}
  \caption{
  Views used for BEV annotation and validation.
  Left: \emph{real} BEV image used to create the Gold manual GT.
  Middle/right: pseudo-aerial BEV reconstructions from SegFormer and diffusion.
  LiDAR returns are overlaid (red) to indicate cells with direct sensor support.
  Annotators label SegFormer/diffusion views independently; the final mask is a fusion that retains only pixels where both views agree on the class.
  }
  \label{fig:pseudo_aerial_views}
  \vspace{-0.8em}
\end{figure}

Each frame is labeled twice---once on the SegFormer reconstruction and once on the diffusion reconstruction---with LiDAR returns overlaid (\autoref{fig:pseudo_aerial_views}).
We fuse both masks using strict agreement with void propagation: a pixel is kept only if both annotations are non-void and assign the same class; otherwise it is void.
Void pixels are excluded from losses and metrics, yielding high-precision supervision on retained pixels at reduced coverage.
\noindent\textit{Protocol validation on Gold (reliability vs.\ coverage).}
We apply the same two-view labeling and fusion on a small set of Gold frames with manual GT (\autoref{fig:pseudo_aerial_views}, left) and evaluate against GT (\autoref{tab:bev_iou_visible_vs_all}).
Static classes achieve consistently high IoU across annotators, while vehicles are harder due to reconstruction artifacts and partial observability.
Fusion yields the highest accuracy (mIoU $=0.896$) but ignores many pixels (Ignored $=0.642$), as disagreements are intentionally converted to void.
Vehicle IoU is similar (and for fusion slightly higher) when restricted to LiDAR-visible vehicles, indicating that most failures occur where sensor evidence is weak.
\noindent\textit{Adding VRUs beyond pseudo-aerial annotation.}
VRU semantics can be added efficiently either via sparse LiDAR fine-tuning (Sec.~\ref{sec:influence_sup_labels}) or via LiDAR point-cloud pseudo-labels (e.g., from a LiDAR-based detector/segmenter) rasterized to BEV, complementing the low-budget pseudo-aerial supervision.

\section{Conclusion}
\label{sec:conclusion}

We presented \textbf{GOLD-BEV}, a framework for learning dense BEV semantic environment maps---including dynamic agents---from ego-centric sensors using time-synchronized aerial imagery as supervision during training. By casting supervision into an aerial-style BEV space, we obtain a natural and efficient interface for dense semantic labeling while avoiding many ambiguities of ego-only BEV targets. Crucially, synchronization between aerial and ground platforms enables learning of moving objects without the temporal artifacts that arise when overhead data are not aligned to the drive. To scale beyond limited aerial coverage, we further learn to synthesize pseudo-aerial BEV imagery from the ego sensors, which supports lightweight human annotation and uncertainty-aware pseudo-labeling to exploit additional unlabeled data.

\section*{Acknowledgment}
The research leading to these results is funded by the German Federal Ministry for Economic Affairs and Energy within the project “NXT GEN AI METHODS – Generative Methoden für Perzeption, Prädiktion und Planung". The authors would like to thank the consortium for the successful cooperation
\bibliographystyle{splncs04}
\bibliography{main}

\clearpage
\appendix

\section{Supplementary Material}
\label{app:supplementary}

\begin{figure*}[ht]
  \centering
  \includegraphics[width=\textwidth]{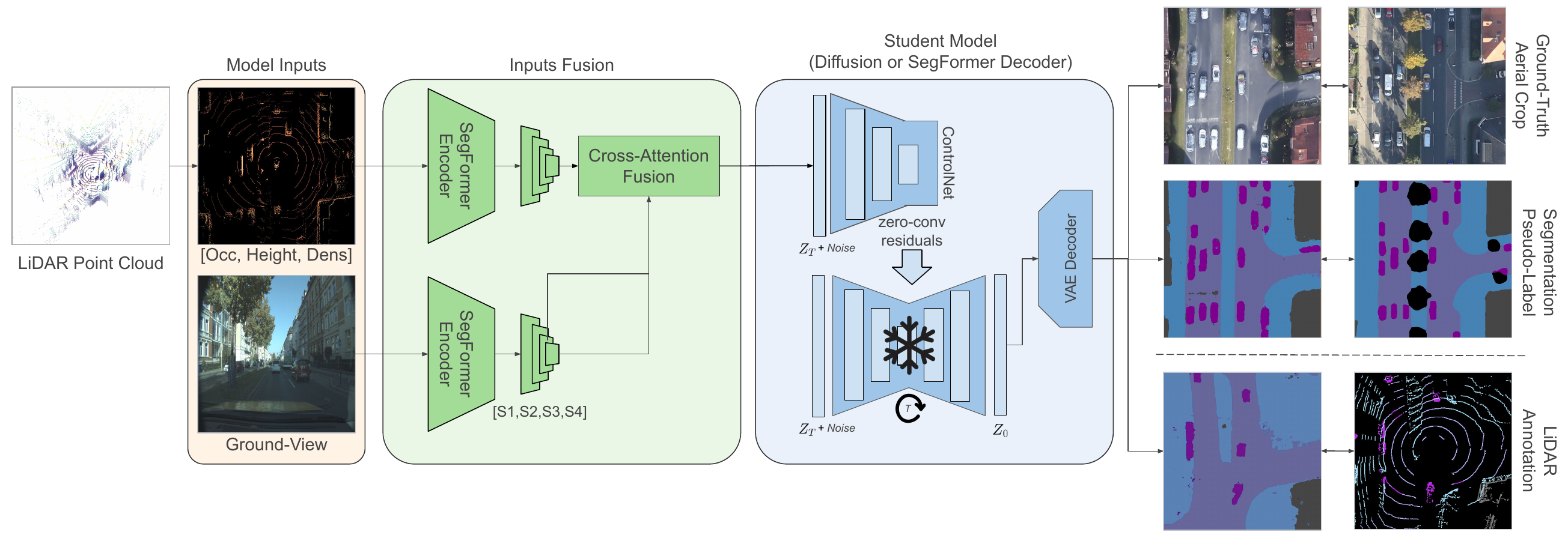} %
  \caption{
Overview of our student model.
The input consists of (i) a LiDAR point cloud rasterized into a BEV representation with occupancy, height, and density channels, and (ii) a front-facing ground-view RGB image.
Each modality is first encoded by a SegFormer backbone.
The resulting multi-scale features are fused by cross-attention, allowing the model to combine complementary information despite the different viewpoints and coordinate systems of camera and LiDAR.
The fused representation is then used as conditioning for a ControlNet-based latent diffusion decoder (or, alternatively, a deterministic SegFormer decoder).
During diffusion, the frozen pretrained denoiser is guided by trainable ControlNet residuals conditioned on the fused features.
After denoising, the final latent is decoded by the VAE decoder to produce a pseudo-aerial reconstruction, as well as semantic segmentation outputs in parallel which can be compared against aerial supervision and LiDAR-based annotations.
}
  \label{fig:arch}
\end{figure*}

\subsection{Network Architectures}

\paragraph{BEV segmentation and reconstruction architecture.}
The model takes as input a forward-facing RGB image and a LiDAR-derived BEV raster.
The RGB image is resized to $600{\times}600$, while the LiDAR point cloud is projected into a BEV grid and represented by three channels encoding occupancy, normalized height, and log-density.
The two modalities are processed by separate SegFormer encoders, each producing four hierarchical feature maps $\{S1,S2,S3,S4\}$.
Because camera and LiDAR features are defined in different coordinate systems, fusion is performed only at the deeper stages $S3$ and $S4$, where the features are more semantic and spatially compact. 

At each fusion stage, the BEV features are projected into a shared latent space and used as queries, while the camera features are projected into the same latent space and used as keys and values.
The feature maps are flattened into token sequences, optionally augmented with positional encodings, and processed by stacked cross-attention blocks.
The attended representation is then projected back to the BEV feature dimension and added to the original BEV feature map through a residual connection followed by normalization.
Thus, the fused representation remains aligned to the BEV stream while incorporating complementary cues from the ground-view image. 

\paragraph{SegFormer decoder branch.}
For the deterministic student, the fused BEV feature hierarchy is decoded by a lightweight SegFormer-style decoder head.
Each stage feature is first projected to a shared hidden dimension, upsampled to a common spatial resolution, concatenated, and fused by a $1{\times}1$ convolution.
A final classifier predicts dense BEV semantic logits at full output resolution.
When BEV reconstruction is enabled, the fused decoder feature is additionally passed through a shallow convolutional reconstruction head to generate a pseudo-aerial BEV RGB image.

\paragraph{The combined latent space as conditioning:.}
For higher-fidelity pseudo-aerial BEV imagery, we additionally train a conditional latent diffusion model based on Stable Diffusion v1.5 with ControlNet.
As illustrated in Fig.~\ref{fig:arch}, the same fused multi-modal representation produced by the two-stream SegFormer encoder and cross-attention fusion serves as the basis for diffusion conditioning.
A lightweight sensor-to-conditioning network converts this fused BEV-aligned representation into a compact control signal, which is provided to the ControlNet branch.
ControlNet then learns trainable residuals that guide the frozen pretrained denoiser, allowing the model to preserve the strong generative prior of Stable Diffusion while grounding the reconstruction in the observed camera--LiDAR input. 

Training follows the standard latent diffusion formulation.
The target pseudo-aerial BEV image is encoded into the latent space of the frozen VAE, noise is added according to the diffusion schedule, and the model is optimized to predict the injected noise.
Supervision is restricted to valid BEV cells, so that invalid or empty regions do not dominate the objective.

\emph{Auxiliary segmentation head.}
To encourage semantic consistency during diffusion training, we optionally attach a lightweight segmentation head to the final feature map of the frozen VAE decoder.
This head predicts BEV semantic logits from the current denoised latent estimate and is supervised jointly with the reconstruction objective on valid BEV cells.

\subsection{Pseudo Labeling}
\begin{figure*}[ht]
  \centering
  \includegraphics[width=\textwidth]{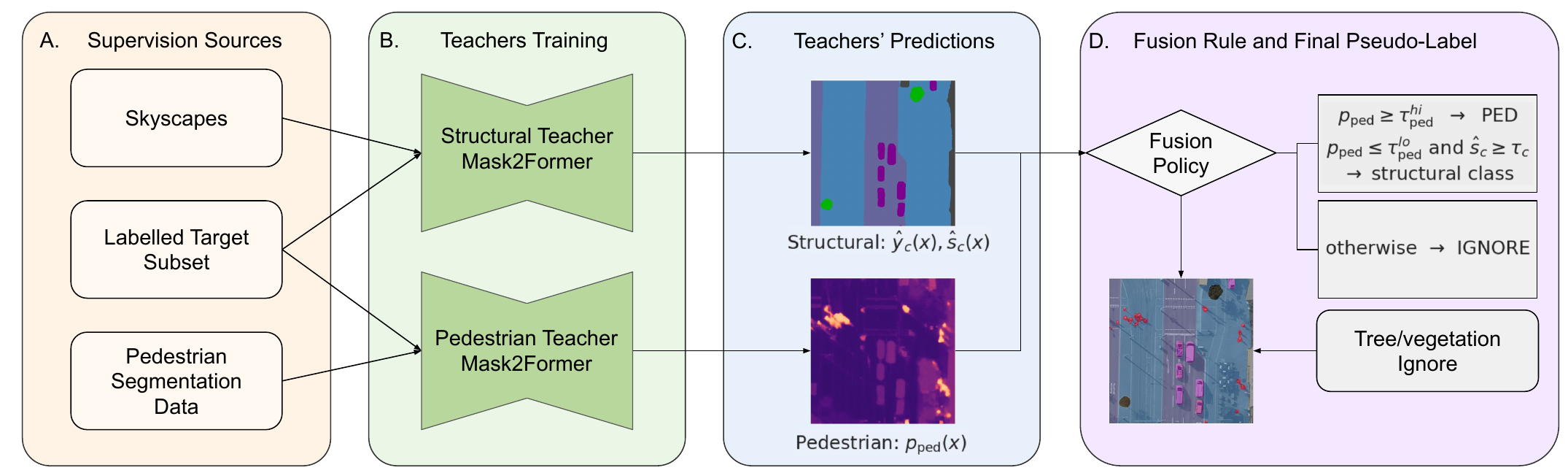} %
  \caption{Starting from two supervision sources—a labeled target subset of BEV-aligned helicopter crops and external supervised data—we train two domain-adapted aerial teachers: a structural Mask2Former for coarse scene layout and a pedestrian-specific Mask2Former for dynamic agents. Applied to each BEV crop, the structural teacher predicts per-pixel semantic labels and confidences, while the pedestrian teacher predicts pedestrian probability. The final dense pseudo-label is obtained by conservative fusion: confident pedestrian predictions overwrite the structural map, confident structural labels are retained where pedestrian evidence is low, and ambiguous pixels are mapped to ignore. Pixels predicted as tree/vegetation are always ignored.}
  \label{fig:pseudoLabel}
  \vspace{-0.8em}
\end{figure*}

Figure~\ref{fig:pseudoLabel} illustrates the pseudo-label generation process described in Sec.~4.2 of the main paper.
A labeled target subset of our aerial crops, the SkyScapes dataset, and an external pedestrian segmentation dataset are used to train two Mask2Former-based aerial teachers: a structural teacher and a pedestrian teacher.
After supervised initialization, both teachers are adapted to the remaining unlabeled target aerial crops using semi-supervised domain adaptation.
For each BEV-aligned aerial crop, the structural teacher produces per-pixel semantic predictions and confidence scores, while the pedestrian teacher outputs a pedestrian probability map.
The final pseudo-label map is obtained by a conservative fusion strategy: confident pedestrian predictions overwrite the structural map, structural labels are retained only when sufficiently confident and not contradicted by the pedestrian teacher, and ambiguous pixels are assigned \texttt{IGNORE}.
In addition, pixels predicted as tree or vegetation are always ignored to avoid supervision errors caused by aerial occlusions.
The resulting fused pseudo-labels provide dense supervision for student training.

\subsection{BEV Image Reconstruction}
Figure~\ref{fig:qual_examples} on the next page provides additional qualitative examples of BEV image reconstruction and BEV semantic segmentation.
A notable observation is the strong consistency between the reconstructed BEV appearance and the predicted semantic layout.
In most cases, structures that are clearly represented in the generated BEV image, such as lane topology, road boundaries, intersections, and highway merge areas, are also reflected coherently in the corresponding segmentation output.
Moreover, the diffusion-based reconstructor benefits from a strong generative prior, which enables it to recover comparatively complex scene geometries with high fidelity, even in challenging scenarios such as large intersections or highway entrances.

\begin{figure*}[ht]
  \centering
  \includegraphics[width=\textwidth]{figures/diff_results.pdf} %
  \caption{Qualitative results on the GOLD-BEV dataset. Each row shows one scene. From left to right: front camera, LiDAR BEV raster, ground-truth BEV aerial crop, reconstructed BEV image, pseudo-label BEV segmentation, and predicted segmentation.}
  \label{fig:qual_examples}
  \vspace{-0.8em}
\end{figure*}

\end{document}